\documentclass[11pt]{article} 
\usepackage{graphicx}

\title{Assembling Actor-based Mind-Maps from Text Streams}
\author{Claudine Brucks and Christoph Schommer \\ University of Luxembourg, Campus Kirchberg.\\
          Dept. of Computer Science and Communication, ILIAS Laboratory\\ 6, Rue Richard Coudenhove-Kalergi, L-1359 Luxembourg\\
         Email: \{ claudine.brucks, christoph.schommer \} @ uni.lu
         }

\begin{document}
\maketitle

\begin{abstract}
For human beings, the processing of text streams of unknown size leads generally to problems because e.g. noise must be selected out, information be tested for its relevance or redundancy, and linguistic phenomenon like ambiguity or the resolution of pronouns be advanced. Putting this into simulation by using an artificial mind-map is a challenge, which offers the gate for a wide field of applications like automatic text summarization or punctual retrieval. In this work we present a framework that is a first step towards an automatic intellect. It aims at assembling a mind-map based on incoming text streams and on a subject-verb-object strategy, having the verb as an interconnection between the adjacent nouns. The mind-map's performance is enriched by a pronoun resolution engine that bases on the work of \cite{STANFORD}.
\end{abstract}

\section{Introduction}\label{secIntroduction}

A text stream is a data flow that is lost once it is read. Such a stream occurs very often in practice, for example while reading a text or listening to a story, probably told by someone else. In both cases, human beings store the major incidents rather associative. First, they remove noise and then extract information out of it, which can either be relevant or redundant/obvious. Then, relevant information is connected very adaptively, meaning that if the same information is read or listened again, the association between co-occurred words increases (or decreases, in case it is not). With such a performance, inconsiderable information gets lost whereas important facts can be kept. This is quite important, because a constructive processing - like the generation of a summarisation of the text and a retrieve of contents - becomes manageable. 

Incremental-adaptive mind-maps serve in a similar way as they simulate such a human performance: through their associative, incremental, and adaptive architecture  they process incoming data streams, adapt internal structures depending on the given input, strengthen or weaken internal connections, and send longer-established connections to a simulating short- and/or long-term memory. In this respect, we base on a work given by \cite{SCHOMMER04} that argues for a real-time approach for finding associative relationships between categorical entities from transactional data streams. Technically, these categorical entities are represented as connectionist cells while associations are represented by links between them. These links may become stronger over time or degrade, according to whether the association re-occurs after a while or not is observed for a while. The work suggests a three-layer architecture: in the first layer, the \emph{short-term memory} treats the incoming signals and constructs the associations. The second layer, which is called the \emph{long-term memory}, stores associations that have a strong connection and that may be useful for a further analysis. The last layer, the \emph{action layer} serves as a communication interface with the user over which he can consult the actual state of the system and interact with it. 

The generation of such a mind-map becomes complicated by the fact that the incoming text can be corrupt or even ambiguous. For example, pronouns produce an ambiguity between existing/referenced persons in the text: having \emph{The President of United States has said that \dots} and a succeeding \emph{Furthermore, he has mentioned that \dots} leads undoubtedly to the same person but the recognition of such relationships is not natural. If we keep such relationships unsolved, the mind-map can become ineffective or even wrong. In this respect, a meaningful part of the intended mind-map described in this work concerns with the resolution of pronouns. For this, we are inspired by some earlier work, notably a syntax based approach \cite{LaL94}. All possible candidates for a pronoun are evaluated on a set of salience factors, as for example recency or subject emphasis. The candidate with the highest salience weight will be chosen as antecedent. \cite{mitkov98robust} presents a similar approach where the candidates are evaluated on indicators, but no syntactic or semantic information on the sentence are needed. Furthermore, the mind-map concerns with a temporal management of text streams to construct an actor-based structure.

\section{Architecture of the Mind-map}

The motivation of pronoun resolution for the semantic network learning is to find the correct antecedent for each pronoun. This is important to construct complete mind-maps for each actor in a text. For this, the text stream is treated by a sliding window, which first buffers and processes a certain number of sentences with the consequence that the information - once it is read - gets lost. For each sentence that is in the sliding window, a pre-defined \emph{subject-verb-object} structure is instantiated and arranged in a semantic network structure, having concepts and connections between them. The connections become stronger or weaker according to the underlying text stream, i.e., the occurrence of the subject-verb-object instantiation.

\begin{figure}[htbp]
   \centering
   \includegraphics[width=7cm]{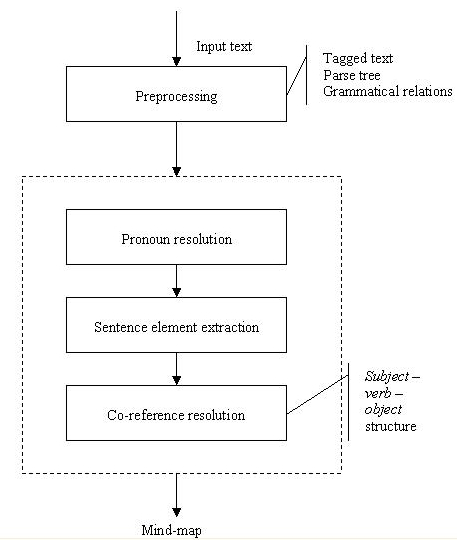} 
   \caption{Architecture}
   \label{fig:architecture}
\end{figure}

Figure \ref{fig:architecture} shows the general architecture of the mind-map. First, the complete text, i.e., each sentence, is preprocessed, which is done in order to get syntactic and semantic information out of the text to further treat the input. In fact, pronouns are used as substitutes for nouns in a text. As an example, the pronoun \emph{he} refers back to \emph{Harry} in a sentence like \textit{Harry goes to the zoo where he looks at the beautiful animals}. Then, a predefined structure of \emph{subject(s), verb(s) and object(s)} is extracted from each sentence as well as the adjacent adjective(s) of both subjects and objects. All these extracted elements are in fact the essence of the sentence. Finally, the co-reference resolution focuses on merging concepts that relate to the same content. As an example, the concepts \emph{President Washington} and \emph{George Washington} relate both to the same person. However, the co-reference resolution is limited to the actors of the text.

\section{Accuracy - Pronoun Resolution}

Following our experiences and looking back at the most important concepts for each category of the text - where \emph{most important} refers to those that have the most outgoing edges - we have observed that these concepts are generally the actors of the stories (this is in respect to stories) whereas for biographies and news articles, the most important concept is the person the biography or news text is about. In scientific texts, the actors are often not the most occurring actors. In respect to the structures that occur multiple times inside a text stream, one can observe that most of all \emph{subject-verb} structures reoccur more often than \emph{subject-verb-object} structures. Those \emph{subject-verb} structures that occur multiple times mostly contain a verb of cognition or communication as for instance: \emph{say}, \emph{think} or \emph{explain}.

In concern of the accuracy of the pronoun resolution - that is how many pronouns are correctly or wrongly resolved and even remain unresolved (see Table \ref{resultPronounRes}) - we have observed that the resolution results applied to pronouns given in third person singular are rather successful. For this, we have used texts from different domains, i.e., fairy tales, news articles, biographies and scientific articles. Only the resolution of \emph{it} and \emph{they} lead to an insufficient accuracy, which demand for an alternative method.

\begin{table}[htbp]
\centering
\begin{tabular}{|l|l|l|l|}
\hline
\textbf{pronoun} & \textbf{correct (\%)} & \textbf{false (\%)} & \textbf{? (\%)}\\
\hline
\textbf{he} & 89.58 & 9.89 & 0.52 \\
\hline
\textbf{his} & 86.11  & 11.80 & 0.69\\
\hline
\textbf{him} & 84.10 & 11.36 & 4.54\\
\hline
\textbf{himself} & 100 & 0 & 0\\
\hline
\textbf{she} & 90.60 & 9.40 & 0\\
\hline
\textbf{her} & 92.24 & 7.76 & 0\\
\hline
\textbf{herself} & 100 & 0 & 0\\
\hline
\textbf{it} & 34.29 & 62.14 & 3.57\\
\hline
\textbf{its} & 44 & 56 & 0\\
\hline
\textbf{itself} & - & - & -\\
\hline
\textbf{they} & 38.84 & 55.37 & 5.79\\
\hline
\textbf{their} & 53.33 & 43.33 & 3.33\\
\hline
\textbf{them} & 60.71 & 37.5 & 1.79\\
\hline
\textbf{themselves} & 100 & 0 & 0\\
\hline
\textbf{I} & 79.71 & 18.84 & 1.45\\
\hline
\textbf{my} & 60 & 40 & 0\\
\hline
\textbf{me} & 75 & 25 & 0\\
\hline
\textbf{myself} & 100 & 0 & 0\\
\hline
\end{tabular}
\caption{Resolving the pronouns: correct, wrong, and unresolved.}
\label{resultPronounRes}
\end{table}

\section{Implementation}
In concern of the implementation, we use a graphical user interface, on which the user can operate, for example to fix the window size, to fix the actors in the text, and to look at the different outputs of the program - as for example the different sub-mind-map related to each actor, diverse actor statistics. For the preprocessing of the text streams, we still need

\begin{itemize}
   \item the tagged text, which permits to filter out all the nouns, proper nouns and pronouns.
   \item the parse tree, which gives more information about the constituents of each sentence, as for example the clauses.
   \item the grammatical relations between the single words of a sentence, relating for example a subject noun with its corresponding verb.
\end{itemize}

\begin{figure}[htbp]
\centering
\includegraphics[width=4cm]{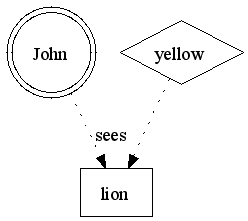}
\caption{Mind-map for \emph{John sees the yellow lion}.}
\label{semGraph}
\end{figure}

\begin{figure}[htbp]
   \centering
   \includegraphics[angle=90, width=11cm]{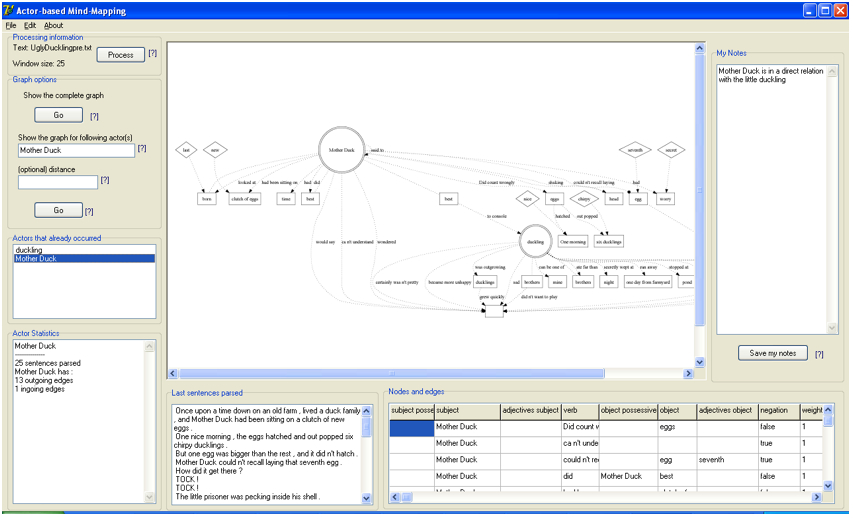} 
   \caption{The workbench.}
   \label{fig:workbench}
\end{figure}

With this, the pronoun resolution works as illustrated in the selective examples:

\begin{itemize}
    \item \emph{he/she}: we take the last male/female noun or name occurring before the pronoun that acts as a subject in the sentence. If there is none, we take the last male/female noun or name before the pronoun.
     \item \emph{they}: we look back at the last two sentences and take the last plural before the pronoun. Plurals remain either plural nouns (e.g. \emph{the women}, \emph{the children}, \emph{the cars}) or noun phrases containing nouns connected by \emph{and} or \emph{,} (e.g. \emph{John and Paul}).
     \item \emph{it}: we detect if \emph{it} is pleonastic or not. If \emph{it} is pleonastic, it has no antecedent as for example in the phrase: \emph{It can be seen that ...}). This is done with the help of a set of some fixed sentence structure patterns (taken from \cite{Dimitrov02alight}). If \emph{it} is not pleonastic, we take the last non-living object occurring before the pronoun which is part of a non-prepositional phrase.
\end{itemize}

To extract the structure of subject-verb-object from each sentence, the grammatical relations described in \cite{STANFORD} are used:

\begin{verbatim}
John sees the yellow lion
\end{verbatim}
with
\begin{verbatim}
nsubj(sees-2, John-1)
det(lion-5, the-3)
amod(lion-5, yellow-4)
dobj(sees-2, lion-5)
\end{verbatim}

The relation \emph{nsubj} (nominal subject) relates the noun \emph{John} with the corresponding verb \emph{sees}, whereas the relation \emph{dobj} (direct object) relates this verb with the object \emph{lion}. In this way, the sentence elements are extracted and the sentence structure can be translated into the mind-map. All subjects and objects take over the roles of the concepts, whereas the verbs serve as connections between the concepts. The adjectives represent sub-concepts of both subjects and objects. From a graphical point of view, actors are represented as double-circles, while concepts that represent no actors are drawn as boxes. The sub-concepts (adjective) are drawn as diamonds. Concepts are linked by a directed arrow, labeled with the verb that relates the subject with the object. An example can be seen in Figure \ref{semGraph}, representing the sentence \emph{John sees the yellow lion}.

In order to merge concepts - that refer to the same actors - we use an incremental actor-based thesaurus. Sine the user can enter different information about the actor - for example the first name, the last name, nicknames, etc. in advance -  we use this external information to establish the thesaurus. Following the spirit of \cite{SCHOMMER04}, the concepts are then matched. Figure \ref{fig:workbench} presents the implemented user-interface consisting of different components, for example the technical (left) part (including processing information, graph options, and actor statistics), the monitoring part (below, including the last parsed sentences and information about each node), and the notes part (to do and save own comments). The workbench is enriched by help buttons.

\section{An Example}

The following text is taken from an extract of the children story \emph{Malcolm the Scotty Dog}. In this example, the focus is on an actor called \emph{Malcolm}. The text is processed sentence-wide. With that, we start with

\begin{verbatim}
Malcolm picked the bone up and ran over 
to the other side of the garden.
\end{verbatim}

\begin{figure}[htbp]
\centering
\includegraphics[width=4cm]{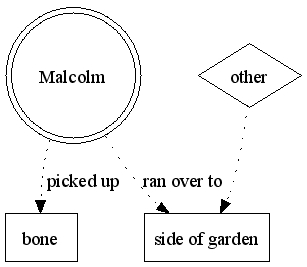}
\caption{Mind-map of \emph{Malcolm} after first sentence}
\label{Malcolm1}
\end{figure}

\begin{figure}[htbp]
\centering
\includegraphics[width=7cm]{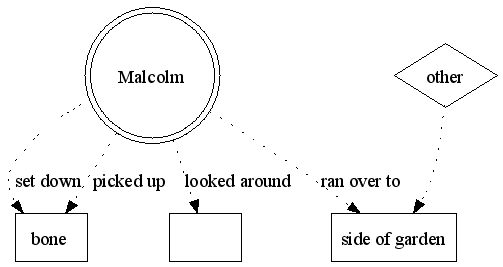}
\caption{Mind-map of \emph{Malcolm} after two sentences}
\label{Malcolm2}
\end{figure}

The mind-map for the actor Malcolm after this sentence can be seen in Figure \ref{Malcolm1}. The actor \emph{Malcolm} is centralized pointing to the concepts  \emph{bone} and  \emph{side of garden}. The last concept is characterised by a sub-concept called \emph{other}.  After the next sentence, the mind-map of \emph{Malcolm} has evolved in the way as represented in Figure \ref{Malcolm2}.

\begin{verbatim}
He set the bone down and looked around.
\end{verbatim}

\begin{figure}[htbp]
\centering
\includegraphics[width=8cm]{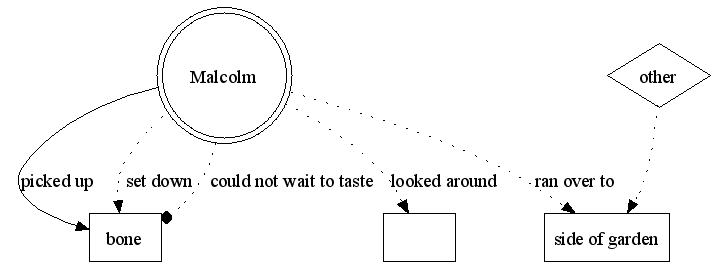}
\caption{Mind-map of \emph{Malcolm} after three sentences}
\label{Malcolm3}
\end{figure}

We observe that \emph{he} has been resolved to \emph{Malcolm}. An empty concept is added since \emph{looked around} does not imply an object. The concept \emph{bone} is stimulated again by \emph{set down} (new concept) and connected to it. With

\begin{verbatim}
He picked it up and could not wait to taste it.
\end{verbatim}

both occurrences of \emph{it} have been replaced by \emph{bone} (Figure \ref{Malcolm3}). The negative verb \emph{could not wait} is specially marked in the mind-map by an \emph{inhibitating} arrow. The phrase \emph{he picked up the bone} has re-occurred in the text stream. To mark this re-occurrence in the mind-map, the structure \emph{Malcolm - picked up - bone} has been enforced (by means of a straight line). Here, it is possible for the reader to display the mind-maps in certain \emph{depths}. By selecting a depth of 1, only the concepts directly related to the actor will be represented, while for a depth of 2, all the concepts at a distance of two nodes will be displayed. This can be illustrated by processing the following sentence.

\begin{verbatim}
The bone was big and it tasted delicious.
\end{verbatim}

\begin{figure}[htbp]
\centering
\includegraphics[width=8cm]{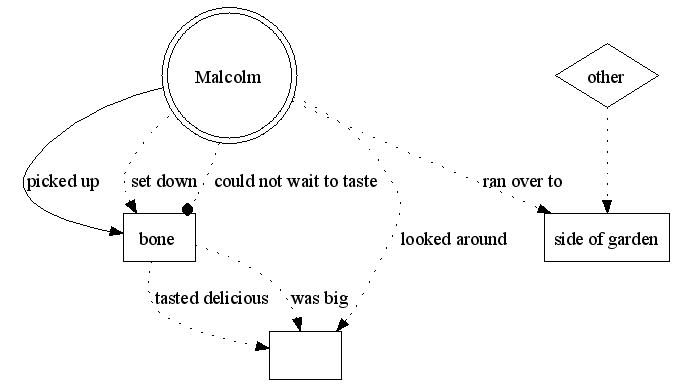}
\caption{Mind-map of \emph{Malcolm} after four sentences, with a depth of 2}
\label{Malcolm4}
\end{figure}

\begin{figure}[htbp]
   \centering
   \includegraphics[angle=90, width=4.2cm]{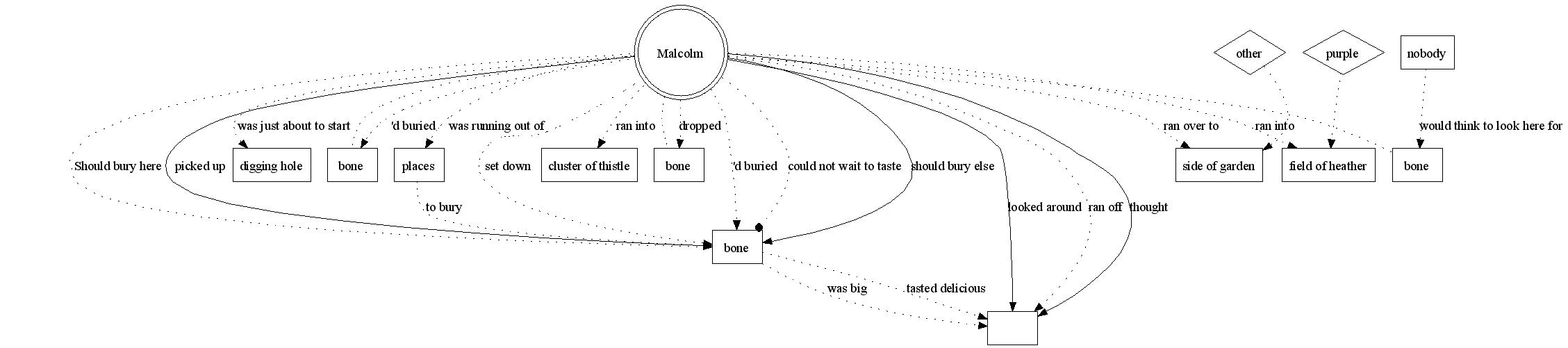} 
   \caption{A Mind-map after having read 20 sentences.}
   \label{fig:mindmap}
\end{figure}

By displaying a depth of 1, the mind-map of \emph{Malcolm} will be as in Figure \ref{Malcolm3}. But when displaying a depth of 2, the mind-map will look as in Figure \ref{Malcolm4}. Here, we notice that the concept \emph{bone} is explained in a more detailed way. And in fact, the user decides how detailed the mind-map should be. Figure \ref{fig:mindmap} shows the mind-map after the processing of a larger amount of sentences.

\section{Conclusions}

The mind-map is a knowledge structure that continuously actualises itself as long as text is read. The representation of the mind-map as a semantic network structure permits to gather all the actions, thoughts and states of being of one actor in a graphical representation. Through the temporal consolidation, changes over time can easily be captured in the mind-map.  Currently, we work on two mind-map extensions that concern with an improved interaction. First, and since a main application is the support of a textual summarization of read text streams, we currently build an automatic text-based summariser. The first (prototypical) version simply outputs the concepts related to an actor, including the sub-concepts and the connections. As the connections are syntactically unchanged, it is easy to generate sentences out of it. Secondly, a selective information retrieval engine is currently done through the extension of the user/mind-map communication through a SQL-like interface. With that, we aim at queries like the following:

\begin{verbatim}
select sub-concepts, concepts from mind-map 
  with depth=1 
  where concept = "Malcolm"
\end{verbatim}

This leads to a result set where all concepts, sub-concepts, and associations are retrieved. The operation \emph{depth} says that only the neighbor elements are considered. In case that depth is set to $\geq 2$, all components of the over-next level are retrieved. A second retrieval then results in a set where only all sub-concepts of \emph{Harry} are retrieved.

\begin{verbatim}
select sub-concepts, name from mind-map
  where name = "Harry"
\end{verbatim}

To be more precise, the following commands are currently under implementation:

\begin{itemize}
   \item \emph{select} : the projection that gives the concepts, sub-concepts, and associations to other concepts.
   \item \emph{from} : the selection to a mind-map; alternatively, several mind-maps can be addressed.
   \item \emph{with depth} : the depth around a concept.
   \item \emph{where} : the where clause allows a diverse condition setting.
\end{itemize}

However, a disadvantage of the mind-map is currently that it grows fast and becomes very large. With this implementation, texts with $\geq 500$ sentences are still an overkill. In this respect, the optimization of the existing solution is a future concern as well. Furthermore, sentences can be composed of not only one single clause, but of several clauses. These clauses are either independent or dependent clauses. Independent clauses can stand as a simple sentence and express a complete thought. Dependent clauses on the other hand can not express a complete thought by standing alone as a sentence. They simply make no sense when standing alone. This is why dependent clauses are connected to an independent clause. This connection is lost in the mind-maps so that some branches going off one actor do not make a lot of sense. Also, the application depends highly on the accuracy of the parser used during the preprocessing. If the parser can not identify the subject(s), verb(s) and object(s) of the sentence, errors or gaps will occur in the mind-maps. Also, as the resolution of some pronouns depends on the correct processing of the sentences, some pronouns may be wrongly resolved due to mistakes of the parser.

\section*{Acknowledgement}

This work has been done within a Master Thesis at the MINE Research Group, ILIAS Laboratory, University of Luxembourg.

{

}

\end{document}